\documentclass[10pt,journal,compsoc]{IEEEtran}

\ifCLASSOPTIONcompsoc
  \usepackage[nocompress]{cite}
\else
  \usepackage{cite}
\fi
\usepackage{multirow}
\ifCLASSINFOpdf
  \usepackage[pdftex]{graphicx}
\else
  \usepackage[dvips]{graphicx}
  \graphicspath{{../eps/}}
  \DeclareGraphicsExtensions{.eps}
\fi
\ifCLASSOPTIONcompsoc \usepackage[caption=false,font=footnotesize,labelfont=sf,textfont=sf]{subfig}
\else \usepackage[caption=false,font=footnotesize]{subfig}
\fi

\begin{document}
%
\title{ Unsupervised Learning in Reservoir Computing for EEG-based Emotion Recognition}
%
%
%
%

\author{Rahma~Fourati,~\IEEEmembership{Student~Member,~IEEE,}
        Boudour~Ammar, ~\IEEEmembership{Senior~Member,~IEEE,}
        Javier~Sanchez-Medina, ~\IEEEmembership{Senior~Member,~IEEE}
        and~Adel M.~Alimi,~\IEEEmembership{Senior~Member,~IEEE}
\IEEEcompsocitemizethanks{\IEEEcompsocthanksitem R. Fourati, B. Ammar and A. M. Alimi are with the Research Groups in Intelligent Machines, Department of Electrical and Computer Engineering, National Engineering School of Sfax, University of Sfax, Sfax 3038, Tunisia\protect\\
E-mail: \{rahma.fourati, boudour.ammar, adel.alimi\}@ieee.org
\IEEEcompsocthanksitem J. J. Sanchez-Medina is with the Innovation Center for the Information Society, University of Las Palmas de Gran Canaria, Las Palmas de Gran Canaria, Spain.\protect\\
E-mail:javier.sanchez.medina@ieee.org}}
\IEEEtitleabstractindextext{%
\begin{abstract}
In real-world applications such as emotion recognition from recorded brain activity, data are captured from electrodes over time. These signals constitute a multidimensional time series. In this paper, Echo State Network (ESN), a recurrent neural network with a great success in time series prediction and classification, is optimized with different neural plasticity rules for classification of emotions based on electroencephalogram (EEG) time series. Actually, the neural plasticity rules are a kind of unsupervised learning adapted for the reservoir, i.e. the hidden layer of ESN. More specifically, an investigation of Oja's rule, BCM rule and gaussian intrinsic plasticity rule was carried out in the context of EEG-based emotion recognition. The study, also, includes a comparison of the offline and online training of the ESN. When testing on the well-known affective benchmark "DEAP dataset" which contains EEG signals from 32 subjects, we find that pretraining ESN with gaussian intrinsic plasticity enhanced the classification accuracy and outperformed the results achieved with an ESN pretrained with synaptic plasticity. Four classification problems were conducted in which the system complexity is increased and the discrimination is more challenging, i.e. inter-subject emotion discrimination. Our proposed method achieves higher performance over the state of the art methods.
\end{abstract}

}

\maketitle

\IEEEdisplaynontitleabstractindextext

%
\IEEEpeerreviewmaketitle

\ifCLASSOPTIONcompsoc
\IEEEraisesectionheading{\section{Introduction}\label{sec:introduction}}
\else
\section{Introduction}
\label{sec:introduction}
\fi

%
%
%
%
\IEEEPARstart{A}{ffective} Computing \cite{poria2017review} is the study and development of systems and devices that can recognize, interpret, process, and simulate human affects. It is an interdisciplinary field spanning computer science, psychology, and cognitive science \cite{tao2005affective}. 
In 1990s, Picard pointed out in her book "Affective Computing" \cite{picard1997affective}, that imbuing machines with the ability of detecting, recognizing, and processing the human emotion is necessary to further enhance human machine interaction. Toward that more reliable interaction, Picard defined three major applications of affective computing: (i) systems that detect and recognize emotions, (ii) systems that express emotions (e.g., avatars, agents), and (iii) systems that feel emotions. Specifically, Emotion Recognition (ER) drew the most attention of scientists. 

In the early stage of affective computing, the proposed works focused on the recognition and synthesis of facial expression, and the synthesis of voice inflection. After that, a variety of physiological measurements are available which would yield clues to one’s hidden affective state.
 Affective wearables offer possibilities of new health and medical research opportunities and applications.  Medical studies could move from measuring controlled situations in laboratory, to measuring more realistic situations in life. 
 
Electroencephalogram (EEG) is a direct measurement of brain activity. EEG gains interest thanks to its excellent temporal resolution, i.e. it is recorded on a millisecond. Note that, human cannot control his brain activity, it is unconscious. In addition, emotion recognition from other modalities which are external manifestations can lead to inaccurate interferences particularly where the included subjects in the experiment control their emotions and suppress them or alter them to exhibit false emotions. Nevertheless, the evoked issue is not available for the case of EEG modality. Emotion recognition based on EEG classifies the inner human emotions. 
The issue for EEG-based emotion recognition comes down to the variability from an individual to another of the recorded EEG signals in response to a stimulus, which remains the inter-subject emotion discrimination a challenging task. 

Diverse writers from psychology defined affect or emotion differently. Ekman \cite{ekman1999basic} distinguished six basic emotions which are happiness, sadness, surprise, disgust, anger, and fear. Whereas, Russel \cite{russell1980circumplex} defined two dimensions of emotion. Although the precise names vary, the two most common categories for the dimensions are “arousal” (calm/excited), and “valence” (negative/positive). Mehrabian \cite{mehrabian1995framework}, \cite{mehrabian1996pleasure} showed that a third dimension is required to discriminate between anger and anxiety. The third dimension tends to be called “dominance”. It ranges from submissive (or without control) to dominant (or in control/empowered). Another interesting representation of emotion belongs to Sun \emph{et al.}. \cite{sun2009improved}. They improved the valence-arousal space in order to handle emotions by computers. The adaptation was improved by the introduction of the typical fuzzy subspace.

One way to design emotion recognition system is the use of neural networks. There are two types of neural networks: feedforward and recurrent. The FeedForward Neural Networks (FFNNs) are characterized by their activation, they fed forward from input to output through "hidden layers" like "Multi-Layer Perceptrons" (MLP), "Radial Basis Function Network" (RBFN) \cite{mellouli2015deep}, etc. However, natural neural networks are usually connected in more complex structures, like recurrent synaptic connections. Recurrent Neural Networks (RNNs) try to emulate that recurrence, which is more suited for dynamical systems modeling.

ESNs are a class of RNN. They are characterized by being three-layered NNs, with a high level of recurrence. The hidden layer is called “reservoir”. ESNs are well fitted for a number of applications, like time series prediction \cite{chouikhi2015hybrid}, \cite{chouikhi2016single}, \cite{chouikhi2017pso}, \cite{slama2017distributed} or robot control \cite{ammar2013learning}, usually showing high performance. As the EEG is temporal signal, ESNs are a natural match for emotion recognition. In our proposed methodology, we used the reservoir to encode the spatio-temporal information of the EEG signals. 
It is well known that the major drawback of the ESN is its random initialization which affects both the convergence state and the performance rate. In the current work, we circumvent this problem by adding an unsupervised learning of the random reservoir before the learning of the output layer. Here, we investigate the reservoir adaptation with two different techniques which are the synaptic plasticity and the intrinsic plasticity (IP). While the formal adapts the weights of reservoir synapses, the latter adapts the intrinsic excitability of each neuron in the reservoir to the emotion recognition task. 

In this paper, we proposed an enhanced ESN for EEG-based emotion recognition adopting the two-dimensional model of emotions. This work is distinguished by neglecting the feature extraction step and feeding ESN directly with the raw EEG as input. To value this choice, we have also extracted the power band features in the time-frequency domain in order to compare the influence of input type on the classification performance. The main contributions of this work are (1) an empirical study of reservoir behaviour improved with unsupervised learning step in an EEG-based emotion recognition task and (2) an extensive validation of the proposed methodology where four classification problems were conducted raising different level of complexity, i.e. Low/High Valence discrimination, Low/High Arousal discrimination, Stress/Calm discrimination and 8 emotional states discrimination. 

The outline of this paper is composed of 4 sections. Section 2 first describes the existing affective benchmarks and next overviews state of the art methods for EEG-based emotion recognition. Section 3 begins with the ESN model, and next introduces the different rules of plasticity and details the proposed approach. Section 4 illustrates the experimental results and discussion. Section 5 summarizes the paper and outlines our future work. 
\section{Literature Review of EEG-based Emotion Recognition}
Several works for EEG-based emotion recognition were proposed including signal processing and machine learning methodologies. In this section, we present the DEAP dataset and the recent works validated using it.
\subsection{Existing Affective benchmarks}
At first blush, there are several works on EEG-based for emotion recognition. Unfortunately, each proposed approach was validated on a specific EEG experiment, i.e. a private EEG dataset. To the best of our knowledge, there are five publicly accessible affective benchmarks: MAHNOB-HCI \cite{soleymani2012multimodal}, DEAP \cite{koelstra2012deap}, SEED \cite{zheng2015investigating}, DREAMER \cite{katsigiannis2018dreamer} and HR-EEG4EMO \cite{becker2017emotion}.
The DEAP dataset is the most used one. 

The elicitation protocol of DEAP dataset used 40 emotional videos. It was conducted on 32 participants (16 male and 16 female) wearing Biosemi Active 2 acquisition system with 32 EEG channels and 8 peripheral sensors. For each trial, participant ratings were expressed in the three-dimensional space such that arousal, valence and dominance. Consequently, varied classification problems were yielded such as Low/High valence (LVHV), Low/High arousal (LAHA) and Low/High Dominance (LDHD). The authors of \cite{garcia2017symbolic} defined two criteria for discrimination of anxiety levels. Stress label is obtained if valence${<}$=3 and Arousal${>}$=5. Calm label is obtained if 4${<}$=Valence${<}$=6 and Arousal${<}$4. Decomposing the valence arousal space into 4 quadrants leads to 4 emotions which are LALV, LAHV, HALV and HAHV \cite{zheng2017identifying}. Combination of the three dimensions leads to 8 emotional states \cite{liu2013eeg} as shown in Table \ref{emotionalstates_table}.
\begin{table}[!t]
\renewcommand{\arraystretch}{1.3}
\caption{Emotional states projection in the VAD space}
\label{emotionalstates_table}
\centering
\begin{tabular}{c||c}
\hline
\bfseries VAD levels & \bfseries Emotional State \\ \hline\hline
HVLALD & Protected\\ \hline
HVLAHD & Satisfied \\ \hline
HVHALD & Surprised \\ \hline
HVHAHD & Happy \\ \hline
LVLALD & Sad \\ \hline
LVLAHD & Unconcerned\\ \hline
LVHALD & Frightened\\ \hline
LVHAHD & Angry\\ \hline
\end{tabular}
\end{table}
\subsection{Related work on DEAP dataset}
\begin{table*}[!t]
\renewcommand{\arraystretch}{1.3}
\caption{Previous works on DEAP dataset}
\label{existingworks_table}
\centering
\begin{tabular}{c||c||c||c||c||c||c||c}
\hline
\bfseries Study & \bfseries Year & \bfseries \parbox{1.6cm}{Subject-Independent} & \bfseries Input& \bfseries \#Channels & \bfseries Classifier & \bfseries Affective states & \bfseries Performance (${\%}$)\\ 
\hline\hline
\cite{pandey2019emotional}&2019&Yes&DWT coefficients&1&MLP&Happy/Sad&58.50\\
\hline\hline
\multirow{2} {*}{\cite{piho2018mutual}}& \multirow{2} {*}{2018} & \multirow{2} {*}{No} & \multirow{2} {*}{Statistical features} & \multirow{2} {*}{32} & \multirow{2} {*}{k-NN} & LAHA & 82.77\\
\cline{7-8}
& & & & & & LVHV & 82.76 \\
\hline\hline
\multirow{8} {*}{\cite{li2018emotion}} & \multirow{8} {*}{2018} & \multirow{8} {*} {Yes} & \multirow{8} {*}{Entropy and energy features of 4s}& 10 & \multirow{8} {*}{k-NN} &
\multirow{4} {*}{LAHA} & {89.81$\pm$0.46} \\
\cline{5-5}\cline{8-8}\
& & & & 14 & &  & {92.24$\pm$0.33} \\
\cline{5-5}\cline{8-8}\
& & & & 18 & &  & {93.69$\pm$0.30} \\
\cline{5-5}\cline{8-8}\
& & & & 32 & &  & {95.69$\pm$0.21} \\
\cline{5-5}\cline{7-8}
 & & & & 10 &  &
 \multirow{4} {*}{LVHV} & {89.54$\pm$0.81} \\
 \cline{5-5}\cline{8-8}\
& & & & 14 & &  & {92.28$\pm$0.62} \\
 \cline{5-5}\cline{8-8}\
& & & & 18 & &  & {93.72$\pm$0.48} \\
 \cline{5-5}\cline{8-8}\
& & & & 32 & &  & {95.70$\pm$0.62} \\
\hline\hline
\cite{garcia2017symbolic}&2017&Yes&Entropy features & 32 & SVM & Stress and Calm & 81.31\\ 
\hline\hline
\multirow{4}{*}{\cite{zhuang2017emotion}} & \multirow{4}{*}{2017} & \multirow{4}{*}{No} & \multirow{4} {*}{IMF features of 5s}& 8 & \multirow{4}{*}{SVM} & \multirow{2}{*}{LAHA} & {71.99} \\
\cline{5-5}\cline{8-8} & & & & 32 & &  & {72.10} \\
\cline{5-5}\cline{7-8}
 & & & & 8 & &\multirow{2}{*}{LVHV} & {69.10}\\
\cline{5-5}\cline{8-8} & & & & 32 & &  & {70.41} \\
\hline\hline
\multirow{3} {*}{\cite{fourati2017optimized}} & \multirow{3}{*}{2017}& \multirow{3}{*}{Yes} & \multirow{3}{*}{Raw EEG} & \multirow{3} {*}{32} & \multirow{3} {*}{ESN} & LAHA & 68.28\\
\cline{7-8}
& & & & & & LVHV & 71.03 \\
\cline{7-8}
& & & & & &8 emotions & 68.79 \\
\hline\hline
\rule {0pt}{15pt} \multirow{2} {*}{\cite{zheng2017identifying}}& \multirow{2}{*}{2017}& \multirow{2}{*}{Yes} & \multirow{2} {*}{Differential entropy of 1s} & \multirow{2}{*}{32} & \multirow{2}{*}{GELM} & \multirow{2}{*}{\parbox{2cm}{LALV, HALV, LAHV, and HAHV}} & \multirow{2} {*}{69.67}\\[17pt] 
\hline\hline 
\multirow{2} {*}{\cite{alhagry2017emotion}} & \multirow{2} {*}{2017} & \multirow{2} {*}{No} & \multirow{2}{*}{Raw EEG segment}& \multirow{2}{*}{32} & \multirow{2}{*}{LSTM} & LAHA & 85.65\\
\cline{7-8}
& & & & & & LVHV & 85.45 \\
\hline\hline
\multirow{4} {*}{\cite{li2015eeg}} & \multirow{4} {*}{2015} & \multirow{4} {*}{No} & \multirow{2} {*}{PSD of 1s}& \multirow{4}{*}{32} & \multirow{4} {*}{SVM} & LAHA & 64.30\\
\cline{7-8}
& & & & & & LVHV & 58.20 \\
\cline{4-4}\cline{7-8}
&  & & \multirow{2} {*}{DBN output from 1s raw EEG}&  &  & LAHA & 64.20\\
\cline{7-8}
& & & & & & LVHV & 58.40 \\
\hline\hline
\multirow{2} {*}{\cite{torres2014comparative}} & \multirow{2}{*}{2014} & \multirow{2} {*}{Yes} & \multirow{2} {*}{Raw EEG}& \multirow{2} {*}{32} & \multirow{2} {*}{HMM} & LAHA & {55.00$\pm$4.5} \\
\cline{7-8}
& & & & & & LVHV & {58.75$\pm$3.8} \\
\hline\hline
\multirow{8}{*}{\cite{wichakam2014evaluation}}& \multirow{8}{*}{2014}& \multirow{8}{*}{Yes} & Bandpower features& 10 &\multirow{8}{*}{SVM}&\multirow{4}{*}{LAHA} & 64.90\\
\cline{5-5}\cline{8-8}
& & & & 32 & &  & {62.90}\\ 
\cline{4-5}\cline{8-8}
& & & PSD features& 10 & & & 63.00\\
\cline{5-5}\cline{8-8}
& & & & 32 & &  & 63.40\\ 
\cline{4-5}\cline{7-8}
& & & Bandpower features & 10 & &\multirow{4}{*}{LVHV} & 64.90\\
\cline{5-5}\cline{8-8}
& & & & 32 & &  & 62.30\\ 
\cline{4-5}\cline{8-8}
& & & PSD features & 10 & & & 56.40\\
\cline{5-5}\cline{8-8}
& & & & 32 & &  & 60.00\\ 
\hline\hline
\multirow{2}{*}{\cite{liu2013eeg}}& \multirow{2}{*}{2013} & \multirow{2}{*}{Yes} & \multirow{2}{*}{Fractal dimension features}& 16 & \multirow{2} {*}{SVM} &
 \multirow{2} {*}{8 emotions} & 65.63 \\
 \cline{5-5}\cline{8-8}
& & & & 32 & &  & 69.53 \\
\hline
\end{tabular}
\end{table*}
In the literature, most of the existing methods begins with the signal decomposition and the feature extraction steps. As EEG signals contain both time and frequency information, feature extraction methods also differ and belongs to three kinds: time-domain, frequency-domain and time-frequency domains. Several works were proposed for classification of 2 valence or arousal levels as in \cite{piho2018mutual}, \cite{li2018emotion}
and \cite{zhuang2017emotion}. There is a lack of proposed works for classification of more than 2 emotions and that's mainly due to the poor achieved results.

A very important aspect in the classifier is whether the task is user dependent or independent. Portioning training data and test data form the same subject and adapting the classifier on a specific subject makes the task subject dependent which has improved results. For instance, in \cite{piho2018mutual} classification of statistical features extracted from DEAP dataset achieves 82.76\% and 82.77\% for 2 valence and arousal levels, respectively with k-Nearest Neighbor (k-NN). But, here the lack of generalization is the cost to pay. Hence, proposing a model trained and tested with data from independent users will lead to an easy application to new subject, i. e. no requirement to design a new model for the new subject.

To achieve this merit, several works focused on the feature extraction step aiming to find the most relevant features for EEG-based ER. Power spectral features (PSD) using Short-Time Fourier Transform as in \cite{li2015eeg} and \cite{wichakam2014evaluation} are considered the baseline features and the most used. Fractal Dimension (FD) \cite{liu2013eeg}, entropy and energy features \cite{li2018emotion} have shown acceptable results as depicted in Table \ref{existingworks_table}. Other works tested features of decomposition techniques such as Discrete Wavelet Transform (DWT) \cite{pandey2019emotional} or Intrinsic Mode Functions (IMF) \cite{zhuang2017emotion}. Most of the work done on EEG-based ER have suffered from finding informative features from EEG data. 

These findings have reshaped scientific understanding of EEG signals and inspired following works to analyze them directly instead of performing the feature extraction step. For example, \cite{torres2014comparative} classified the EEG preprocessed signals using the Hidden Markov Model (HMM). 
Likewise, feature learning was performed by feeding the raw channel data to the Deep Belief Network (DBN) \cite{li2015eeg}. The new representation obtained from DBN is then fed to Support Vector Machine (SVM). A comparison with PSD features showed that raw EEG data achieved better results. The task here is subject-dependent and the length of the trial is 1 second.
Note that, EEG signals are acquired from a number of channels. 

Investigation of the impact of specific channels on the ER performance is very important. \cite{pandey2019emotional} finds that one channel F4 is sufficient for an EEG-ER task. Also, the authors in \cite{wichakam2014evaluation} showed that bandpower features from 10 channels handle better results than using 32 channels for classifying 2 levels of valence and arousal, respectively.
In the current work, ESN has as input the preprocessed EEG raw and the classification is also performed by the readout layer. 

The use of ESN for processing EEG data is not new at all, but the works that are based on ESNs are really few. In fact, there are only two works \cite{bozhkov2016learning} and \cite{bozhkov2017reservoir}. We should notice that the input is event-related potential features. Bozkhov \emph{et al.} \cite{bozhkov2016learning} have pretrained the reservoir with IP in order to have optimal values of IP parameters, i.e. the gain and the bias. The next step is to compute the steady states of the reservoir. Bozkhov \emph{et al.} \cite{bozhkov2016learning} proved that the discrimination of the steady states is more efficient due to the adequate representation of the input data with the IP. Therefore, the best performance 76.9\% found with using Linear Discriminant Analysis (LDA). 

Bozkhov \emph{et al.} have also integrated a feature selection step by using projection of 2D, 3D and 4D of the steady states. The input of the ESN is a feature vector with 252 attributes. The experiment was conducted on 26 females seeing positive and negatives emotional pictures from IAPS and wearing an EEG cap with 21 channels. The best recognition rate up to 98.1\% was by using 4D projections with SVM. In a follow up work of Bozkhov \emph{et al.} \cite{bozhkov2017reservoir}, it was demonstrated that the data representation with reservoir computing outperforms the autoencoder. In fact, the highest classification accuracy achieved is 81\% with 2 layers autoencoders.

Recently, \cite{fourati2017optimized} proposed ESN pretrained with IP and fed it with preprocessed EEG data. The aim of the work is to show the effectiveness of IP rule on the reservoir layer for performing feature extraction step. The classification of arousal levels, valence levels and 8 emotional states resulted in 68.28\%, 71.03\% and 68.79\% respectively.

Thus, the use of ESN and its promising results in ER encourage us to further use the ESN for feature extraction and classification of emotions from EEG signals. We are different from state-of-the-art methods \cite{bozhkov2016learning} and \cite{bozhkov2017reservoir}. We use ESN as an architecture for both representation of input data and its classification. The projection and the classification ranking in \cite{bozhkov2016learning} and \cite{bozhkov2017reservoir} are more complex and hence the computational time is high.

While our work focus only on the EEG modality, we emphasize a recently published work using ESN pretrained with IP having as input both EEG and physiological signals\cite{ren2018emotion}. After calculating the asymmetry index from the frontal lobe channels, and based on its values some signals were discarded in the recognition process. Next, Wavelet Packet Transform (WPT) is applied to extract 4 sub-bands. After that, k-means is used to cluster the generated WPT coefficients of each window in each channel. This work used ESN for feature selection step that means to reduce the size of the first feature vector. The output of the reservoir states is then fed to SVM with RBF kernel. The authors in \cite{ren2018emotion} tested the proposed methodology on DEAP dataset labeled with the 4 quadrants LALV, HALV, LAHV and HAHV to yield 78.2\% as recognition rate. Compared to our work, our feature extraction step is more simple and benefit from all information of the raw EEG data. 

Our objective goes more than one form of plasticity, we further investigate a second form and two techniques of ESN training which are the offline and the online modes. Even if in \cite{yusoff2016modeling} the authors studied the effect of synaptic plasticity with these two modes, it looks that the study was evaluated on time series of small size and the IP form was ignored. Briefly, we distinguish our work by illuminating the important roles of both adapting the reservoir and training mode to a more challenging task, i. e. EEG-ER.

\section{Methodology}
In our methodology, we consider the preprocessed EEG signals. To justify such choice, we also perform a feature extraction step and fed power bands features to ESN to recognize emotions. Note that, our methodology consists in three steps which are feature extraction, reservoir pretraining with one kind of plasticity rule and readout layer training for the classification.
\subsection{Feature Extraction}
\begin{figure}[!t]
\centering
\includegraphics[width=3.5in,height=1.9in]{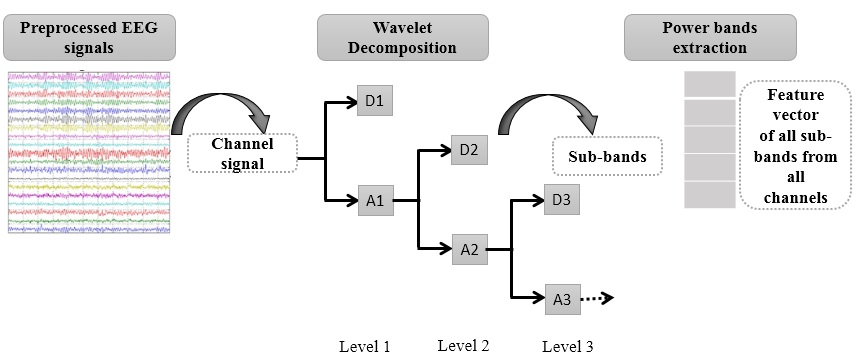}
\caption {Power bands feature extraction process.}
\label{fig_fe}
\end{figure}

\begin{figure*}[!t]
\centering
\subfloat[]{\includegraphics[width=3.5in]{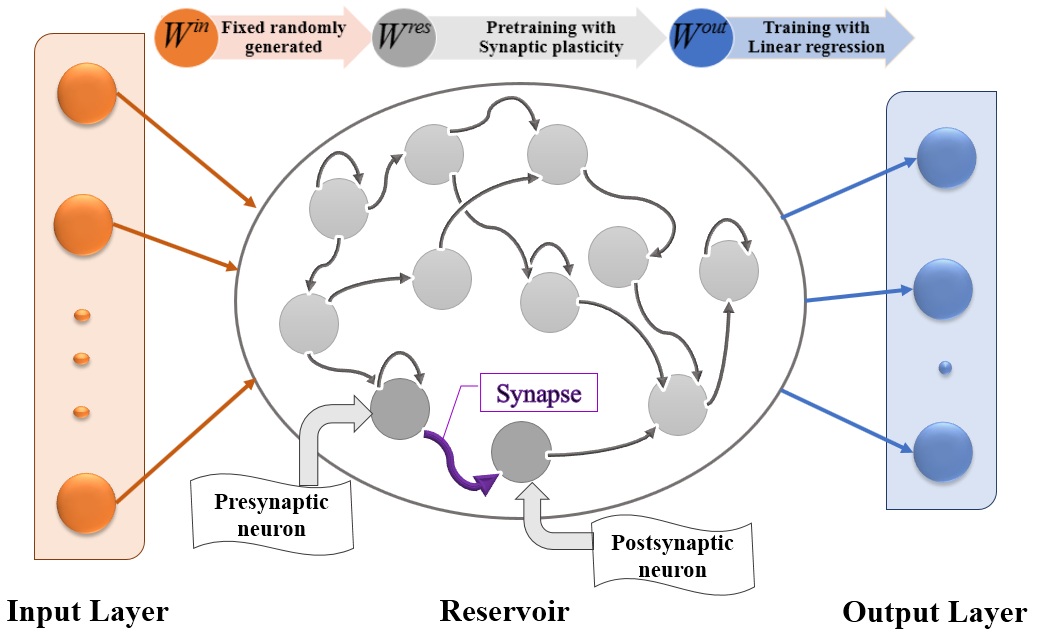}
\label{fig_first_case}}
\hfil
\subfloat[]{\includegraphics[width=3.5in]{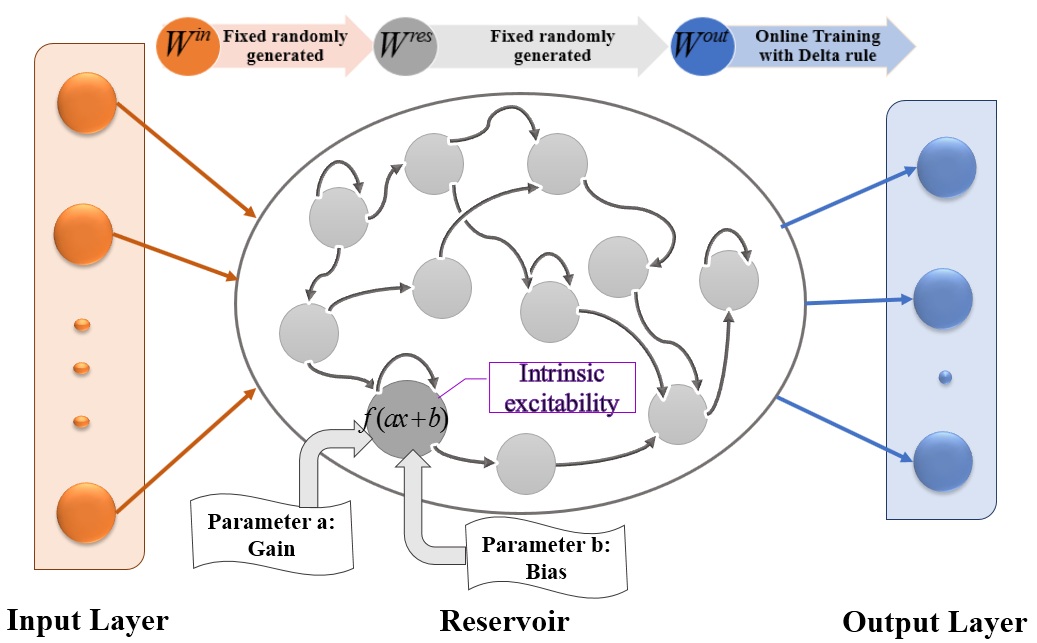}
\label{fig_second_case}}
\caption{Unsupervised learning of Reservoir layer using (a) Synaptic plasticity rule and offline mode and (b) Intrinsic plasticity rule and online mode.}
\label{reservoirlearning}
\end{figure*}
EEG signals are not stationary. To handle this hypothesis, it was shown that wavelet decomposition is more convenient than Fourier transform \cite{chaabane2017wavelet}. In our study, we applied DWT to decompose EEG channel signal into different bands. The DWT coefficients are defined as follows: 
\begin{equation}
\label{dwt_eqn1}
C_{_{x(t)}}(l,n)=\int_{-\infty }^{\infty}x(t)\Psi _{l,n}(t)dt
\end{equation}
\begin{equation}
\label{dwt_eqn2}
\Psi _{l,n}(t)=2^{-(l+1)}\Psi(2^{-(l+1)}(t-2^{-1}n))
\end{equation}
The EEG signal, $x(t)$, is correlated with a wavelet function $\Psi _{l,n}(t)$. The variable $l$ and $n$ are the scale and translation variables of the wavelet function, respectively. 
A dyadic scale is performed as in (\ref{dwt_eqn2}) to chose the scale and translation variables. Note that, the original signal is reconstructed if orthogonality is ensured \cite{addison2017illustrated}.
The scale variable provides analysis in frequency domain: compressed version of the wavelet function corresponds to the high frequency components of the original signal while the stretched version corresponds to the low frequency components; and the translation variable provides analysis in the time domain. 
The decomposition of DWT leads to coefficients of high frequency which are "detail" and coefficients of coarse approximation of original signal in time domain which are "approximation".
After decomposition, power features from all bands are extracted. Note that, power bands features are the most popular features in the context of EEG-based emotion recognition. The Definition of EEG frequency bands differs slightly between studies. Finally, a feature vector is formed through the concatenation of all power features from all channels of the same EEG trial.
Thus, the size of the feature vector is equal to the number of the bands multiplied by the number of channels. Fig. \ref{fig_fe} illustrates the feature extraction process adapted in our methodology.   

\subsection{ESN Model}
ESNs are a type of recurrent neural networks proposed by Jaeger \cite{jaeger2002tutorial}. It is composed of input layer, a reservoir and an output layer often called readout layer as depicted in Fig. \ref{fig_first_case}. The recurrence in ESNs is handled through the recurrent connections between hidden  units and the possible feedback connection  from the output layer to the reservoir. Direct connection from input to output layer can be added. 

The basic structure proposed for solving computational intelligence problems was the FFNN. However, for specific problems which have dynamic and temporal nature, FFNNs are not able to handle complex temporal machine learning problems. As a solution, recurrent connections are added to the structure, giving birth to RNNs for handling several problems such as SVM leaning process \cite{fourati2015improved}. Training RNNs was done by extending the canonical backpropagation algorithm to be Back-Propagation-Through-Time (BPTT) \cite{werbos1990backpropagation}. One of the major limitations of BBTT is its high computational cost and its slow convergence. Hence, Jaeger \emph{et al.} \cite{jaeger2002tutorial} proposed the reservoir computing approach to enhance the learning process. 

The reservoir is composed of sparsely connected neurons with cycled connections. The weights from input to reservoir and the inner weights of reservoir are randomly initialized and remain unchanged during training phase. The simplicity of ESN turns out in the need of training only the weights of the readout layer which is a linear regression. 
Consider a topology of $I$ input neurons, $R$ internal neurons and $O$ output neurons. The first step in training ESN is to collect the matrix of activation states of each neuron in the reservoir. The activation equation at each time step is expressed in (\ref{activation_eqn}):
\begin{equation}
\label{activation_eqn}
x(t)=f_{res}\left ( W^{in}u(t)+W^{res}x(t-1) \right )
\end{equation}
where $W^{in}$  and $W^{res}$ are the weights of the input and the reservoir layers respectively. $f_{res}$  is the non-linear activation function, usually, a sigmoid.
The second step consists in calculating the output weights. Here, we distinguish two modes: offline and online. Linear regression is calculated as follows in (\ref{wout_eqn}) and it is considered as an offline mode:
\begin{equation}
\label{wout_eqn}
W^{out}=Y_{target}*\left ( X^{T}X \right )^{-1}
\end{equation}

The training in the online mode is ensured through the minimization between the target output and the produced output. The presentation of the training samples is in the sequential form.

Originally proposed for MLP model, the delta rule is a stochastic gradient descent method used for the update of one layer of MLP neural network \cite{pemberton1988generalized}. It is expressed as follows:
\begin{equation}
\label{deltawout_eqn}
\Delta W=\eta \left ( y^{desired} (t)-y(t)\right )x(t)
\end{equation}
\begin{equation}
\label{woutonline_eqn}
W^{out}=W^{out}+\Delta W
\end{equation}
where $\eta$ is the learning rate and t is the time step of the learning iterations, $t = 1, 2, … , T$. $x(t)$ is the vector of neuron firing activation states of $x$ at time step $t$. This mechanism computes the incremental adaptation of readout weights.

Then, the output of the ESN can be generated according to (\ref{readtout_eqn}) using linear regression if $f_{out}$ is a linear function:
\begin{equation}
\label{readtout_eqn}
y(t)=f^{out}\left ( W^{out}*x(t) \right )
\end{equation}
For a specific input sample, the neuron having the highest activation score wins, while the other neurons in the same layer are inhibited. This mechanism is called winner-take-all.
Assuming that a trial is composed of $N$ channels, the ESN is fed sequentially with these $N$ signals one by one. As a matter of fact, the ESN output is a channel label. Each successive $N$ channels labels are combined  through  majority  voting  and  the  class label receiving the most number of votes is regarded as the trial label.
\subsection{Plasticity Rules for Unsupervised Learning of Reservoir Layer}
The generation of fixed random reservoir makes the RNN training fast. Meanwhile, other studies in neuroscience have reported that the modification of the connection strength endows neural networks with a powerful learning ability. This mechanism is called Synaptic plasticity illustrated in Fig. \ref{fig_first_case}. Here we detail the most used existing ones, to know, the Oja's rule and the BCM rule. Recent findings showed that the neuron is able to change its intrinsic excitability to fit the distribution of the input as shown in Fig. \ref{fig_second_case}. We also explain the gaussian intrinsic plasticity mechanism.
\subsubsection{Oja's rule}
Oja learning rule, proposed by Erkki Oja \cite{oja1982simplified}, is a model of how neurons in the brain or in artificial neural networks alter the strength of connections, or learn, over time. Oja's rule is an extension of hebbian learning rule \cite{hebb1994organization} proposed in Hebb book "The Organization of Behaviour". In classical hebbian learning rule the update scheme for weights may result in very large weights when the number of iterations is large.

Oja's rule is based on normalized weights, the weights are normally normalized to unit length. This simple change results in a totally different but more general and stable weight update scheme compared to classical hebbian learning scheme.
It can also be extended to non-linear neurons as well. It has been mathematically proven that when the input data is centered at the origin and when Oja's update rule converges it results in the neuron learning the first principal component of the training data, hence Oja's rule is important in feature based recognition systems \cite{oja1982simplified}.

The Oja learning rule can be described as follows in (\ref{oja_eqn}):
\begin{equation}
\label{oja_eqn}
\Delta W_{kj}(t)=\xi y_{k}(t)\left [ x_{j}(t)-y_{k}(t)W_{kj}(t) \right ])
\end{equation}
where  $W_{kj}$ is the change of the synaptic weight between the postsynaptic neuron, $y_{k}$, and the presynaptic neuron,$x_{j}$, at time, t. $\xi$ is the learning rate. Note that Eq. (\ref{oja_eqn}) is a modified version of the anti-Hebbian rule by adding a forgetting factor to limit the growth of the synaptic weight to avoid the saturation of $W_{kj}$ .
\subsubsection{BCM rule}
The BCM rule \cite{castellani1999solutions}, named for Bienenstock, Cooper and Munro, follows the Hebbian learning principle, with a sliding threshold as a stabilizer function to control the synaptic alteration. 

The main idea was that the sign of weight modification should be  based  on  whether  the  postsynaptic  response  is  above  or  below  a  threshold. Responses  above  the  threshold  should  lead  to  strengthening  of  the  active  synapses, responses below the threshold lead to weakening of the active synapses. 
The threshold varies  as  a  nonlinear function of the average output of the postsynaptic neuron, which is the main concept of  the  present  BCM  model providing stability  properties.

The BCM rule has several variants and here the one suggested in \cite{castellani1999solutions} is adopted as given by (\ref{bcm_eqn1}) and (\ref{bcm_eqn2}):
\begin{equation}
\label{bcm_eqn1}
\Delta W_{kj}(t)=y_{k}(y_{k}-\theta _{M})x_{j}/\theta _{M}
\end{equation}
\begin{equation}
\label{bcm_eqn2}
\theta _{M}=E\left [ y_{k}^{2} \right]=\sum p_{k}y_{k}^{2}
\end{equation}
where $\theta _{M}$  is the modification threshold of the postsynaptic neuron, $y_{k}$, and $p_{k}$ is the probability of choosing vector of  from the dataset, $E\left[  \right]$ is the temporal average, $W_{kj}$  is the adjustment of the synaptic weight between postsynaptic neuron, $y_{k}$ and presynaptic neuron, $x_{j}$, at time t.
\subsubsection{Intrinsic plasticity rule}
From a biological point of view, Triesch \cite{triesch2005gradient} stated that the biological neuron does not adapt its synapses, rather it adapts its intrinsic excitability. While traditional learning algorithms update the weights of connections between neurons, the IP rule algorithm update the activation function of the neuron.
Particularly, Triesch derived the IP rule for fermi activation function and for an exponential desired distribution. In the same manner, Schrauwen \emph{et al.} \cite{schrauwen2008improving} extended IP considering also hyperbolic tangent as activation function with Gaussian desired distribution. 
 
IP rule is local, that means it is applied on each single neuron to maximize information about its input. From information theory, entropy measure allows us to realize the maximization of information.
Equation (\ref{ip_eqn1}) measures the distance between the actual probability density of the neuron’s output and the targeted probability density using the Kullback-Leiber divergence metric.

\begin{equation}
\label{ip_eqn1}
D_{KL}\left ( p(x),p_{d}(x)) \right )=\int p(x)log\left ( \frac{p(x))}{p_{d}(x)} \right )
\end{equation}
The Kullback-Leiber divergence can be developed into (\ref{ip_eqn2}) for a Gaussian distribution with a mean $\mu$ and a standard deviation $\sigma$.
\setlength{\arraycolsep}{0.0em}
\begin{eqnarray}
\label{ip_eqn2}
D_{KL}\left ( p(x),p_{d}(x)) \right )&{}={}&-H(x)+\frac{1}{2\sigma ^{2}}E\left ( \left ( x-\mu \right )^{2} \right )\nonumber\\ 
&&{+}\:log\frac{1}{\sigma \sqrt{2\pi} }
\end{eqnarray}
\setlength{\arraycolsep}{5pt}

A balance is achieved between the maximization of the actual entropy $H$ and the minimization of the expected entropy $E$. The update of the gain $a$ and bias $b$ is handled using
(\ref{ip_eqn3}) and (\ref{ip_eqn4}). 
\begin{equation}
\label{ip_eqn3}
\Delta a=\frac{\eta}{a}+\Delta b\left ( W^{in}u+W^{res}x \right )
\end{equation}
\begin{equation}
\label{ip_eqn4}
\Delta b=-\eta \left ( -\frac{\mu }{\sigma ^{2}}+\frac{x}{\sigma ^{2}}\left ( 2\sigma ^{2}+1-x^{2}+\mu x \right ) \right )
\end{equation}
Each neuron in the reservoir layer will be activated using (\ref{ip_eqn5}).
\begin{equation}
\label{ip_eqn5}
x(t)=f_{res} \left( {diag}(a)\left( W^{in}u(t)+W^{res}x(t-1)\right) +b \right)
\end{equation}

Schrauwen \emph{et al.} \cite{schrauwen2008improving} showed that updating IP parameters in a gaussian distribution with hyperbolic tangent activation function is similar to the update in an exponential distribution with fermi activation function.
Hence, there is no dependence between the chosen non-linearity function and the targeted distribution.

A very interesting unsupervised rule, the IP, is able to make reservoir computing more robust in a fashion that its internal dynamics can autonomously tune themselves independently of the randomly generated weights or the scaling of input to the optimal dynamic regime for emotion recognition.
For reservoir enhancing purpose, we follow previous works in which they used the IP rule \cite{triesch2005gradient}, \cite{schrauwen2008improving} and \cite{steil2007online}, \cite{wardermann2007intrinsic}.
\section{Experimental Results and Discussion}
In this section, we present and discuss our results using different configurations. To validate our method, we tested it on DEAP benchmark in order to be able to compare our results to the current state-of-the-art methods. Details of experimental settings are provided. An interpretation of emotion recognition results is done. Finally, a sensitivity analysis of ESN to the hyperparameters is detailed. 
\subsection{Experimental Settings}
\begin{table}[!t]
\renewcommand{\arraystretch}{1.3}
\caption{Wavelet Decomposition OF EEG channel signal}
\label{wavelet_table}
\centering
\begin{tabular}{c||c||c}
\hline
\bfseries Bandwith (Hz) & \bfseries Frequency Band & \bfseries Decomposition level\\
\hline\hline
64-128 Hz & Noise & D1\\
\hline
32-64 Hz & Gamma & D2\\
\hline
16-32 Hz & Beta & D3\\
\hline
8-16 Hz & Alpha & D4\\
\hline
4-8 Hz & Theta & D5\\
\hline
1-4 Hz & Delta & A5\\
\hline
\end{tabular}
\end{table}
For the feature extraction step, we chose daubechies function ‘db5’ with 5 levels for wavelet decomposition. We choose 5 levels because the sampling rate of recorded EEG signals in DEAP dataset is 128Hz. Besides, a filter pass band [4-45 Hz] was done when preprocessing DEAP dataset, therefore the delta band was not considered in our work. Power features are generated from 4 bands using the 32 channels resulting in a feature vector with 128 as size. Table \ref{wavelet_table} shows details about frequency bands and corresponding wavelet levels.
To train and test our model, we used 80\% as a training partition and the remaining 20\% as a test partition.
\subsection{Emotion Recognition Results}
In this subsection, we present the results of 4 classification problems using DEAP dataset.
\begin{table}[!t]
\renewcommand{\arraystretch}{1.3}
\caption{Arousal Discrimination Results}
\label{arousal_table}
\centering
\begin{tabular}{p{2cm}||c||c||c||c}
\hline
\bfseries System & \bfseries Input type & \bfseries Offline & \bfseries Online& \bfseries Hybrid\\
\hline\hline
\multirow{2} {*} {ESN-Oja rule} & Feature & 59.77\% & 49.22\%&60.01\%\\ \cline{2-5}
 & Signal & 54.30\% &58.98\% &\textbf{60.29}\%\\
\hline\hline
\multirow{2} {*} {ESN-BCM rule} & Feature & 61.72\% & 54.14\%& \textbf{62.17\%}\\ \cline{2-5}
 & Signal &56.25\% &50.00\% &59.34\%\\
\hline\hline
\multirow{2} {*} {ESN-IP rule} & Feature & 61.21\% & 59.11\%&62.39\%\\ \cline{2-5}
 & Signal & 68.28\% \cite{fourati2017optimized}& 62.98\% &\textbf{69.23\%}\\
\hline
\hline
SVM with PSD features \cite{wichakam2014evaluation}& Feature&\multicolumn{3}{c}{63.40\%}\\
\hline
\hline
 HMM \cite{torres2014comparative} & Signal & \multicolumn{3}{c}{55.00$\pm$4.5\%}\\
 \hline
\end{tabular}
\end{table}

\begin{table}[!t]
\renewcommand{\arraystretch}{1.3}
\caption{Valence Discrimination Results}
\label{valence_table}
\centering
\begin{tabular}{p{2cm}||c||c||c||c}
\hline
\bfseries System & \bfseries Input type & \bfseries Offline & \bfseries Online & \bfseries Hybrid\\
\hline\hline
\multirow{2} {*} {ESN-Oja rule} & Feature &59.77\% &52.73\%& 60.81\%\\ \cline{2-5}
 & Signal & 61.26\% &54.92\%& \textbf{62.13\%}\\
\hline\hline
\multirow{2} {*}{ESN-BCM rule} & Feature &57.42\% &46.88\% &58.26\%\\
\cline{2-5}
 & Signal & 56.25\% & 41.67\%&\textbf{59.31\%}\\
\hline\hline
\multirow{2} {*} {ESN-IP rule} & Feature & 53.52\% &55.86\% &57.94\%\\
\cline{2-5}
 & Signal & 71.03\% \cite{fourati2017optimized} & 66.23\%& \textbf{71.25\%}\\
\hline
\hline
SVM with bandpower features \cite{wichakam2014evaluation}& Feature&\multicolumn{3}{c}{62.30\%}\\
\hline
\hline
 HMM \cite{torres2014comparative} & Signal & \multicolumn{3}{c}{58.75$\pm$3.8\%}\\
\hline
\end{tabular}
\end{table}
\begin{table}[!t]
\renewcommand{\arraystretch}{1.3}
\caption{Emotional states Discrimination Results}
\label{emostates_table}
\centering
\begin{tabular}{p{2cm}||c||c||c|c}
\hline
\bfseries System & \bfseries Input type & \bfseries Offline & \bfseries Online & \bfseries Hybrid\\
\hline\hline
\multirow{2} {*}{ESN-Oja rule} & Feature & 35.49\% & 42.23\% &48.29\%\\\cline{2-5}
 & Signal & 54.29\% & 58.12 &\textbf{59.29\%}\\
\hline\hline
\multirow{2} {*} {ESN-BCM rule} & Feature & 32.42\% & 33.98\%& 41.58\%\\
\cline{2-5}
 & Signal & 56.69\% & 59.81\% &\textbf{60.23\%}\\
\hline\hline
\multirow{2} {*} {ESN-IP rule} & Feature &38.22\%  &44.65\% & 49.58\%\\
\cline{2-5}
 & Signal & 68.79\% \cite{fourati2017optimized}&69.25\% & \textbf{69.95\%}\\
\hline\hline
SVM with FD features\cite{liu2013eeg}&Feature&\multicolumn{3}{c}{69.53\%}\\
\hline
\end{tabular}
\end{table}

\begin{table}[!t]
\renewcommand{\arraystretch}{1.3}
\caption{Stress/Calm Discrimination Results}
\label{anxiety_table}
\centering
\begin{tabular}{p{2.25cm}||c||c||c||c}
\hline
\bfseries System & \bfseries Input type & \bfseries Offline & \bfseries Online &\bfseries Hybrid\\
\hline\hline
\multirow{2} {*} {ESN-Oja rule} & Feature &  65.45\%  & 47.27\%&60.36\%\\\cline{2-5}
 & Signal& 61.82\%  & 65.45\% &\textbf{67.27\%}\\
\hline\hline
\multirow{2} {*}{ESN-BCM rule} & Feature & \textbf{65.45\%} & 50.91\%&64.63\%\\
\cline{2-5}
 & Signal &  49.09\% & 54.55\%&58.29\%\\
\hline\hline
\multirow{2} {*}{ESN-IP rule} & Feature &  69.06\% & 65.45\%& \textbf{76.15\%}\\
\cline{2-5}
 & Signal & 41.82\% & 49.09\%&61.27\%\\
\hline
\hline
 SVM with entropy features\cite{garcia2017symbolic}
& Feature & \multicolumn{3}{c}{81.31\%}\\
\hline
\end{tabular}
\end{table}
For arousal classification, using ESN with offline training which is the linear regression yields better results than the online training using the delta rule. An exception is made for the case of the reservoir pretrained with the Oja's rule using the online training of output weights as depicted in Table \ref{arousal_table}. The comparison is handled with the work in \cite{torres2014comparative} since it belongs to the same context as the current work. 

We highlight that using feature as input achieves higher results while using BCM rule. But, the best accuracy achieved is 69.23\% with ESN pretrained with IP using the hybrid mode inputted directly with EEG channel signal. 
Note that, training the output weights in a first step using linear regression and using the delta rule as a second step represents the hybrid mode. The idea here is that instead using the online mode with random initial weights, we trained them first with linear regression. As a result, the performance increased up to 14\% over the state of the art method \cite{torres2014comparative}.

For valence classification, the reservoir pretrained with IP using the hybrid training reaches the highest result up to 71.25\%. Our system, again, outperforms the existing work using HMM with signal as input \cite{torres2014comparative} up to 13\% as shown in Table \ref{valence_table}.
Furthermore, Oja's rule achieves higher result with signal as input 62.13\% than the feature 60.81\%. While, the BCM rule achieves higher performance 59.31\% with signal than the feature input 58.26\%.

The discrimination of 8 emotional states is considered as the most challenging task in our work since the complexity of the system is increased. 
The online mode is more efficient here than the offline mode for both signal and feature input. The hybridization allows us to achieve the best result which is 69.95\%. In the literature, most of the existing works classify arousal and valence levels. There is only one work which classifies 8 emotional states \cite{liu2013eeg}. Our ESN trained with linear regression followed with the delta rule outperforms the SVM classifier with FD features.

According to Table \ref{emostates_table}, the ESN is more robust when using signal instead of feature. 
When discriminating stress and calm, we remark that ESN with synaptic plasticity achieves the same accuracy with signal as well feature input which is 65.45\%. Combining linear regression and delta rule training enhanced the performance from 61.82\% to 67.27\%. 
While using ESN with IP, we can rise the accuracy to 76.15\% with bandpower features. When inputting signal to ESN with IP, we obtained an accuracy of 41.82\% and 49.09\% with the offline mode and online mode, respectively. The hybrid mode shows its effectiveness in this case to reach 61.27\% as recognition rate.

We believe that existing work combining SVM classifier with entropy features achieves the best result up to 81.31\%, but we highlight that we proved that feeding ESN with EEG signal directly can achieve better result than bandpower features which is the aim of our work as depicted in Table \ref{anxiety_table}. 

Overall, the online mode is better than the offline mode for the classification of the emotional states. But, for the other problems the offline outperforms the online mode. We can conclude that the delta rule is more efficient when the complexity of the architecture is high than the linear regression.
Pretraining the reservoir with Synaptic plasticity rules achieves higher results using the feature vector as input than the raw EEG signal. The only case the synaptic plasticity reaches good accuracy rate with signal is the classification of stress / calm states using Oja's rule.

In all cases, pretraining ESN using intrinsic plasticity rule has achieved the best results either with using feature as input or raw EEG signal. As a consequence, we recommend the use of IP rule and the hybrid mode for classification of EEG signals.
\subsection{Impact of Plasticity Parameters}
\begin{figure}[!t]
\centering
\includegraphics[width=3.5in]{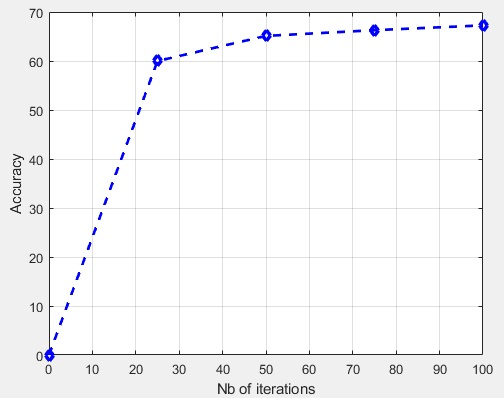}
\caption {Impact of Oja's parameter on the performance.}
\label{fig_ojaimpact}
\end{figure}
Through the aforementioned results, we deduced that in three cases the IP rule achieves better results that SP rules, i. e. arousal, valence and 8 emotional states discrimination. But, for the discrimination of stress and calm states the Oja's rule achieves the highest result.

In our study, we delve into the plasticity parameter which is the number of iterations.
Our results showed that the previous findings are the same for our case for the IP rule. After 10 iterations, the IP is stable and adding more iterations does not improve the result. 
Nevertheless, performing more iterations up to 100 of Oja's rule increases the recognition rate as shown in Fig.\ref{fig_ojaimpact}.

\subsection{Impact of spectral radius and reservoir size}
\begin{figure}[!t]
\centering
\includegraphics[width=3.5in]{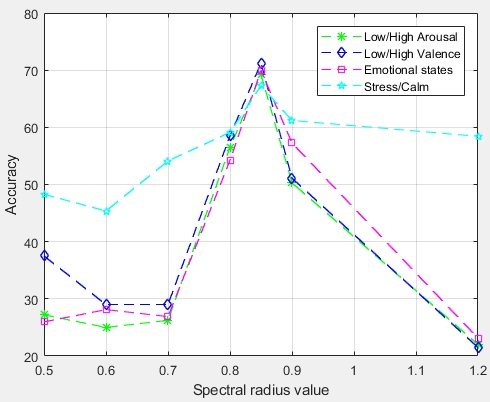}
\caption {Impact of spectral radius on the performance.}
\label{fig_specimpact}
\end{figure}
One of the most important parameters in the ESN model is the spectral radius. We recall that Jaeger \cite{jaeger2002tutorial} recommended that the reservoir should satisfy the ESP. It relates asymptotic properties of the excited reservoir dynamics to the driving signal. 

Intuitively, the ESP states that the reservoir will asymptotically wash out any information from initial conditions. The ESP is granted for any input if this spectral radius is smaller than unity. In Fig. \ref{fig_specimpact}, it is shown that the best value of spectral radius is 0.85 for our classification problems.
\begin{figure}[!t]
\centering
\includegraphics[width=3.5in]{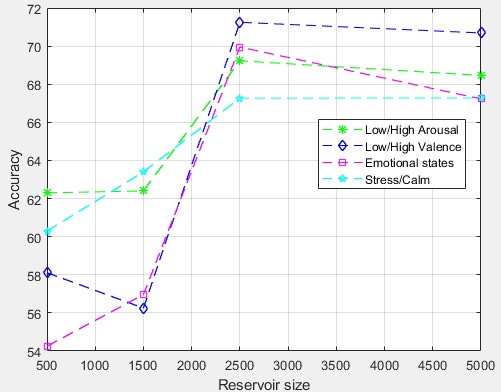}
\caption {Impact of reservoir size on the performance.}
\label{fig_reservoirimpact}
\end{figure}
In order to evaluate the sensitivity of our proposed system, we also varied the reservoir parameter from 500 to 5000 neurons as depicted in \ref{fig_reservoirimpact}. The choice was not arbitrary, rather we tried to take into consideration the channel data length which is 8064. The role of the reservoir here is to compress the input data and to provide a new representation for classification purpose.

The smaller size 500 does not result in a good representation of channel data leading to low accuracy results.
For a medium size 1500, the performance is acceptable for both emotional states and stress/calm discrimination. But, for arousal and valence discrimination the performance is very low.
Using 2500 as a reservoir size enhances the performance for classification problems. That means the reservoir is able to make a new representation of the raw EEG signal making the new feature vector more discriminative.
When increasing the reservoir size to 5000 neurons, the performance is slightly lower. This is mainly due to the increased complexity of the system. It can also include that the reservoir is stable with 2500 neurons.

\section{Conclusion}
This paper presented the prominent computational models of plasticity and their current applicability to empirical improvements as neural network adaptation mechanisms. The presented forms of plasticity are applied to randomly connected recurrent reservoirs to learn the structural information in EEG signals, achieve sparsity in neural connectivity and enhance learning performance of emotion recognition.
We advocate the use of intrinsic plasticity for reservoir pretraining since it achieves best results in comparison with synaptic plasticity for EEG signals classification. We suggest that synergestic learning of the reservoir could yield ESN model with benefits from each paradigm. We are expecting that these suggestions will result in a more robust reservoir computing approach for EEG-based emotion recognition.


%



\ifCLASSOPTIONcompsoc
  \section*{Acknowledgments}
\else
  \section*{Acknowledgment}
\fi

The research leading to these results has received funding from the Ministry of Higher Education and Scientific Research of Tunisia under the grant agreement number LR11ES48.

\ifCLASSOPTIONcaptionsoff
  \newpage
\fi



%
\bibliographystyle{IEEEtran}
\bibliography{IEEEabrv,IEEEexample}{}


%
\begin{IEEEbiography}
[{\includegraphics[width=1in,height=1.25in,clip,keepaspectratio]{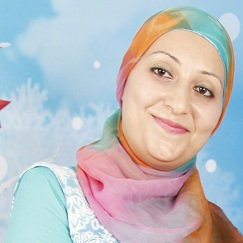}}]{Rahma Fourati}
(IEEE Graduated Student Member ’15). She was born in Sfax, she is a PhD student in Computer Systems Engineering at the National Engineering School of Sfax (ENIS), since October 2015. She received the M.S. degree in 2011 from the Faculty of Economic Sciences and Management of Sfax (FSEGS). She is currently a member of the REsearch Group in Intelligent Machines (REGIM).Her research interests include Recurrent Neural Networks, Affective computing, EEG signals analysis, Support Vector Machines, Simulink Modeling. 
\end{IEEEbiography}

\begin{IEEEbiography}
[{\includegraphics[width=1in,height=1.25in,clip,keepaspectratio]{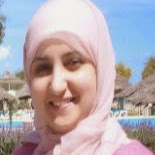}}]{Boudour Ammar}
(IEEE Student Member08, Member13, Senior Member'17) was born in Sfax, Tunisia. She graduated in computer science in 2005. She received the Master degree in automatic and industrial computing from the National School of Engineers of Sfax, University of Sfax
in 2006. She obtained a PhD degree in recurrent neural network learning model for a biped walking simulator with the Research Group on Intelligent Machines (REGIM), University of Sfax,
since February 2014. She is currently an assistant professor with the Department of Computer Engineering and Applied Mathematics at National Engineering School of Sfax (ENIS). Her research interests include iBrain (Artificial neural networks, Machine learning, Recurrent neural Network) and i-health (Autonomous Robots, Intelligent Control, Embedded Systems, medical applications, EEG, ECG).
\end{IEEEbiography}

\begin{IEEEbiography}
[{\includegraphics[width=1in,height=1.25in,clip,keepaspectratio]{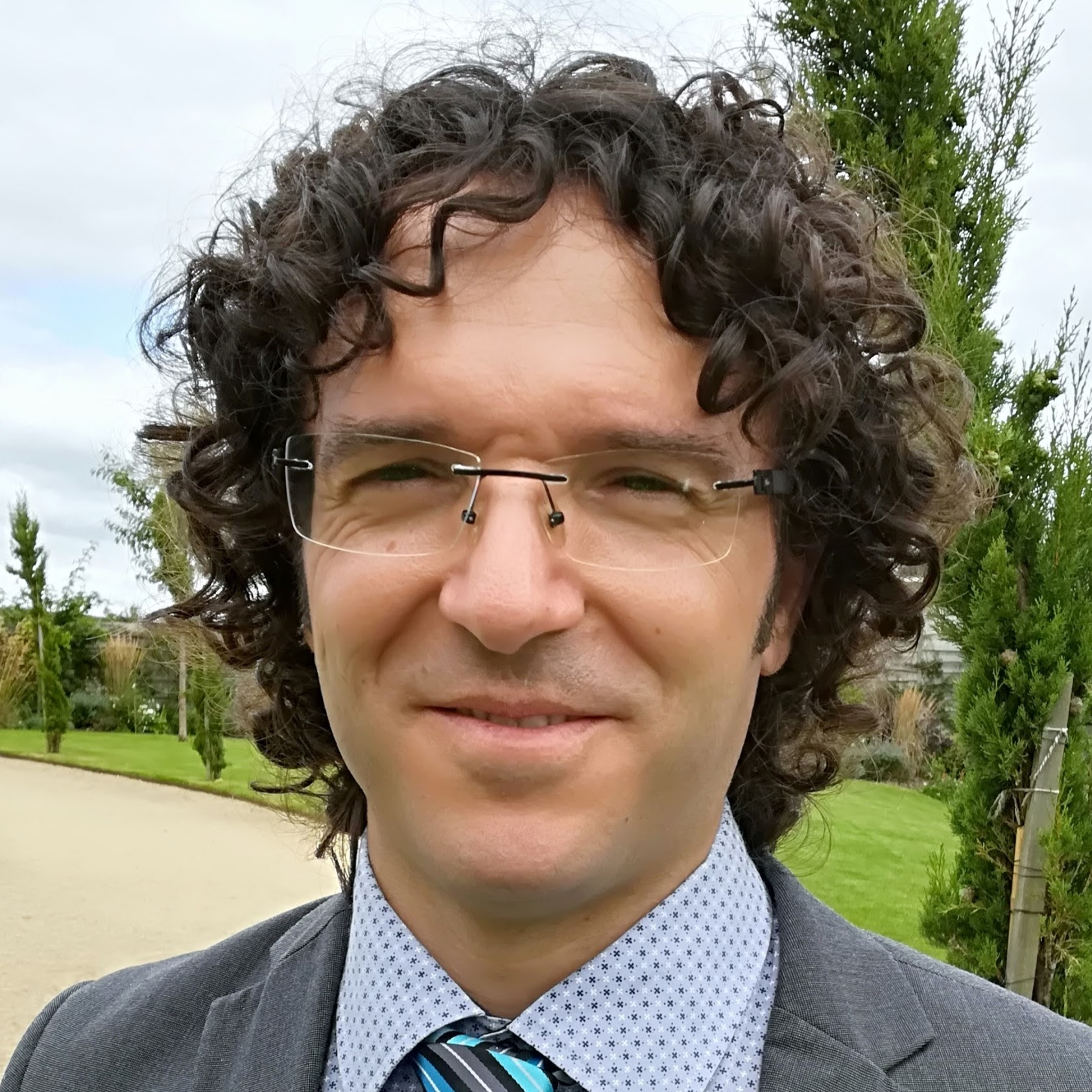}}]{Javier Sanchez-Medina}
(M’13) received the M.E. degree from Telecommunications Faculty in 2002 and the Ph.D. degree from the Computer Science Department, Universidad de Las Palmas de Gran Canaria, in 2008. He has authored over 30 international conference and 15 international journal papers. He is interested in the application of evolutionary and parallel computing techniques for intelligent transportation systems. He is also very active as a Volunteer of the IEEE ITS Society, where he has been serving in a number of different positions. He is the Editor-in-Chief of the ITS Podcast, the ITS Newsletter Vice President of the IEEE ITSS’s Spanish chapter, and the General Chair for the IEEE ITSC2015.
\end{IEEEbiography}
\begin{IEEEbiography}
[{\includegraphics[width=1in,height=1.25in,clip,keepaspectratio]{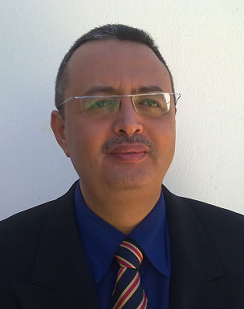}}]{Adel M. Alimi}
(IEEE Student Member’91, Member’96, Senior Member’00).He was born in Sfax (Tunisia) in 1966. He graduated in Electrical Engineering 1990, obtained a Ph.D. and then an HDR both in Electrical and Computer Engineering in 1995 and 2000 respectively. He is now a Professor in Electrical and Computer Engineering at the University of Sfax. His research interest includes applications of intelligent methods (neural networks, fuzzy logic, evolutionary algorithms) to pattern recognition, robotic systems, vision systems, and industrial processes. He focuses his research on intelligent pattern recognition, learning, analysis and intelligent control of large scale complex systems. He is an Associate Editor and Member of the editorial board of many international scientific journals (e.g. Pattern Recognition Letters, Neurocomputing, Neural Processing Letters, International Journal of Image and Graphics,
Neural Computing and Applications, International Journal of Robotics and Automation, International Journal of Systems Science, etc.). He was a Guest Editor of several special issues of international journals (e.g. Fuzzy Sets and Systems, Soft Computing, Journal of Decision Systems, Integrated Computer Aided Engineering, Systems Analysis Modeling and Simulations). He is an IEEE senior member and member of IAPR, INNS and PRS
\end{IEEEbiography}




\end{document}